\documentclass{article}
\pdfoutput=1
\usepackage{graphicx}
\usepackage{amssymb}
\usepackage{multirow}
\usepackage{amsmath}
\usepackage[utf8]{inputenc} 
\usepackage[T1]{fontenc}    
\usepackage{hyperref}       
\usepackage{url}            
\usepackage{booktabs}       
\usepackage{amsfonts}       
\usepackage{nicefrac}       
\usepackage{microtype}      
\usepackage{xcolor}         
\usepackage[ruled,vlined]{algorithm2e}
\usepackage{color,soul}
\usepackage[numbers]{natbib}

\newcommand{\src}{G}
\newcommand{\tar}{F}
\newcommand{\res}{\hat{F}}
\newcommand{\srcdomain}{\mathcal{V}}
\newcommand{\tardomain}{\mathcal{U}}
\newcommand{\mapping}{\Phi_F}

\newcommand{\brwonian}{\mathbf{w}}
\newcommand{\scorenet}{s_\theta}
\newcommand{\score}{\dfrac{\partial \log  p(x)}{\partial x} }

\newcommand{\dataa}{\textit{Gold Atlas dataset}}
\newcommand{\datab}{\textit{CuRIOUS dataset}}
\newcommand{\datac}{\textit{IXI dataset}}

\usepackage[]{hyperref}

\begin{document}
\newtheorem{theo}{Theorem}
\newtheorem{proof}{Proof}
\newtheorem{prop}{Proposition}
\newtheorem{defi}{Definition}
\newtheorem{lemma}{Lemma}
\title{Zero-shot-Learning Cross-Modality Data Translation Through Mutual Information Guided Stochastic Diffusion}

\author{Zihao Wang$^1$\footnote{corresponding: zihao.wang@ieee.org}, Yingyu Yang$^2$, \\Maxime Sermesant$^2$, Herv\'{e} Delingette$^2$, Ona Wu$^1$\\
$^1$ A. A. Martinos Center for Biomedical Imaging@MGH,\\ Harvard University.\\
149 13th St, Charlestown, MA 02129\\
$^2$ Centre Inria
d'Université Côte d'Azur. \\
2004 Rte des Lucioles, 06902 Valbonne, France\\
{\tt\small zwang63@mgh.harvard.edu}
}
\maketitle

\begin{abstract}
Cross-modality data translation has attracted great interest in image computing. Deep generative models (\textit{e.g.}, GANs) show performance improvement in tackling those problems. Nevertheless, as a fundamental challenge in image translation, the problem of Zero-shot-Learning Cross-Modality Data Translation with fidelity remains unanswered. This paper proposes a new unsupervised zero-shot-learning method named Mutual Information guided Diffusion cross-modality data translation Model (MIDiffusion), which learns to translate the unseen source data to the target domain. The MIDiffusion leverages a score-matching-based generative model, which learns the prior knowledge in the target domain. We propose a differentiable local-wise-MI-Layer ($LMI$) for conditioning the iterative denoising sampling. The $LMI$ captures the identical cross-modality features in the statistical domain for the diffusion guidance; thus, our method does not require retraining when the source domain is changed, as it does not rely on any direct mapping between the source and target domains. This advantage is critical for applying cross-modality data translation methods in practice, as a reasonable amount of source domain dataset is not always available for supervised training.
We empirically show the advanced performance of MIDiffusion in comparison with an influential group of generative models, including adversarial-based and other score-matching-based models.
\end{abstract}

\section{Introduction}
One often wants to use existing resources to solve new problems without having to develop similar resources for new data modalities to maximize economic utility. For example, software systems to perform a specific task, e.g. brain parcellation, may be designed to work with a particular data modality, e.g. MRI. Ideally, one would be able to translate the new data modality, e.g. CT, into the specific data modality and be able to analyze the new data with the existing software. This method is often termed cross-modality data translation, which has been of particular interest to the generative learning research community in the past few years. Two major types of methods have been proposed: (1) generative adversarial-based translation and (2) mapping-based translation. Those two approaches correspond to the two major machine learning model types: generative and discriminative learning. The mapping-based method directly learns the mapping between different modalities through regression between the source and target modal data points.
\begin{figure}
    \centering
    \includegraphics[width=0.89\textwidth]{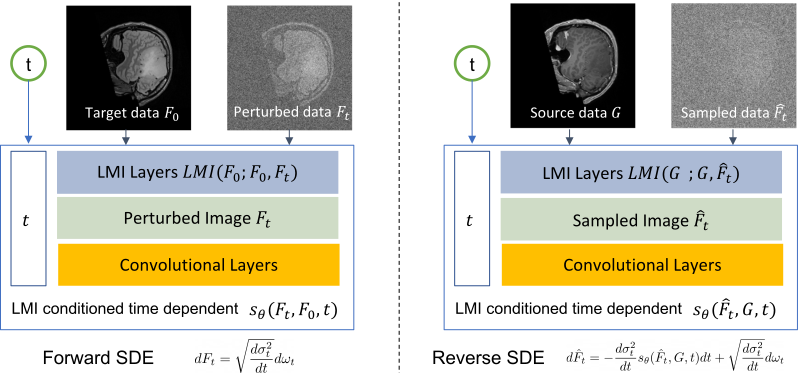}
    \caption{The proposed local-wise mutual information (LMI) guidance for zero-shot stochastic diffusion-based modality translation. Unlike the classical score-matching model, the proposed method conditions on a statistical measure: local-wise mutual information. It measures the LMI between target image $\tar$ and its perturbed image $\tar_t$ at time $t$ during training time (left sub-figure) for the forward SDE. At test time (right sub-figure), it calculates the LMI between source image $\src$ and the sampled image $\res_t$ at time $t$ for the reverse SDE.}
    \label{fig:overview}
\end{figure}
However, paired datasets of different modalities are not always available in many scenarios. Also, slight pixel-wise misalignments between different modalities may lead to inaccurate translation because mapping-based translation models are designed to build a direct mapping between the source and target modalities. These issues have led to recent research focused on generative adversarial-based translation, which learns the cross-modality mapping through cycle-consistency training, which does not require paired datasets. 

Two conceptual metrics have been proposed to evaluate the translation quality: synthesis realism  (see Def. \ref{def;real}) and faithfulness (see Def. \ref{def;faith})\cite{sdeit}. Translation faithfulness refers to how well the translation results maintain the context of the source data intact. Translation realism means how consistent the translation results are with the range of the target modality. Ensuring those two metrics in a Zero-shot-Learning way is challenging as a Zero-shot-trained model does not have any information from the source domain. 

Mapping-based methods\cite{van1994grey, ZhangSAR, mr2ct, han2017mr,nie2016estimating} usually favor translation faithfulness over realism. Many early mapping-based modality translations methods rely on pixel-level modeling between the source and target modality. Generative adversarial network (GAN) \cite{gan1, gan2, gan3, gan4, gan5, gan6, cmtgan7} based methods were proposed due to various shortcomings of the previous mapping-based approaches. The generative model is usually used for modeling the target modality directly, thus achieving translation realism\cite{gan8,gan9, gan7}. This method usually involves complicated adversarial architecture design and modality task-specific loss function design for different translation tasks \cite{sdeit}.
Although the generative adversarial-based translation does not need access to paired datasets for training, it still requires source domain data, which may be challenging to collect, leading to insufficient samples to balance the cycle-consistency training.
Recent works \cite{dhariwal2021diffusion, diff_cond1,diff_con2,nonuni} show that score-based generative models achieve better generation performance than that GAN-based models. Meng \textit{et. al.} \cite{sdeit} proposed SDEdit, which employs a Stochastic Diffusion Model (SDM) \cite{songsde} to perform image translation that balanced faithfulness and realism in the way of Zero-shot-Learning. Different from conventional GAN or mapping-based models, which are nearly end-to-end generation, an SDM is a score-based generative model \cite{DDPM} that relies on iteratively denoising a diffusion sequence driven by an SDE. 
Given a datapoint from the source domain, SDEdit perturbs the datapoint with Gaussian noise (perturbation-based guidance). Then a reverse SDE is used by SDEdit to gradually project the disturbed datapoint to the manifold of the target domain.
SDEdit has been shown to be superior to GAN-based translation models in terms of parsimony of model structure and complexity of loss function. 
Although SDEdit overcomes the drawbacks of previous GAN-based translation methods, it remains limited in cross-modality data translation applications. The perturbation-based guidance assumes the source and target domains can be perturbed by noise, which may not be true in specific translation tasks (e.g., microscopy brightfield $\rightarrow$ darkfield translation). In addition, SDEdit requires optimizing initial time $t_0$ to find the best interval for perturbation.\cite{sdeit}
Muzaffer \textit{et. al.} \cite{GANDIFF} proposed SynDiff to tackle this issue by introducing a cycle-consistent architecture devised with a bilateral diffusion process. The advantage of SynDiff is that the transformation is semantically consistent, yet the computational cost of computing a bilateral diffusion process is doubled. In fact, although it is claimed to be an unsupervised method, SynDiff needs to pre-train a generator to estimate a paired source image for training purposes, which requires an additional generation module; thus, the model's overall performance is also influenced by the pre-training quality of the generator.

We propose a new zero-shot unsupervised learning-based cross-modality data translation method based on a stochastic diffusion process. Instead of seeking any conditioning guidance from the data domain, our model leverages the statistical feature-wise homogeneity for conditioning the diffusion process (see Fig. \ref{fig:overview}). This allows us to bridge the source and target domain by using their local-wise statistical attributions for cross-modality data translation.
The proposed method overcomes the shortcomings of current modality translation approaches:

(1) Unlike GAN-based strategies, the proposed method (MIDiffusion) neither needs inversion optimization in the test step nor requires additional adversarial architectures and loss functions design (cycle consistency, \textit{etc.}). 

(2) In contrast to the current SDEdit method, our framework does not require the optimization of the hyper-parameter $t_0$, which is a user-selected parameter, to achieve balance between realism and faithfulness.\cite{sdeit} 

(3) MIDiffusion is a fully unsupervised Zero-shot-Learning model that can translate  unseen modalities into target modalities.

(4) MIDiffusion also empirically shows that the training of an additional generator \cite{GANDIFF} for diffusion process guidance is not necessary to achieve semantic consistency.

\section{Related Works}
\textbf{Generative Learning-based Cross-modality Data Translation}
A plethora of works performs cross-modality data translation through conditional generative adversarial networks (cGANs) \cite{cmtgan1, cmtgan2, cmtgan3, cmtgan5, cmtgan6, cmtgan7}. Those cGANs-based cross-modality data translation usually requires the generation of a supervised signal for conditioning the generator network (e.g., Zhuge \textit{et. al.} \cite{cmtgan7} used a register to generate the conditional signal).
Some other GANs-based cross-modality data translation methods use cycle consistency training to swap the features between different domains \cite{cmtgan4, cycleganmt1, cycleganmt2, cycleganmt3}.

\textbf{Kernel-based Generative Learning} 
The proposed method falls in the scope of transition-kernel-based generative learning \cite{SongGen,SalimansGen,flowwang, Implicit, hvae, LevyGen}; specifically, it belongs to score-based generative learning \cite{classifierfree, DDPM,score1, score2}. The score-based generative models show comparable data modeling performance to those generative adversarial methods \cite{gan1, gan2, gan3,gan4,gan5,gan6,gan7,cmtgan7, JiangGAN}.

\textbf{Zero-shot Learning}
The cross-modality data translation problem without access to the source modal data leads to a Zero-shot-Learning-based cross-modality data translation problem.
One challenge of learning-based methods is that their modeling ability is restricted when dealing with unseen data classes \cite{wangfewshot,bucher2017generating, zsl}. Zero-shot-Learning is a robust learning scheme to deal with such cases when training and test classes are disjoint. One traditional approach of Zero-shot-Learning tries to find a direct projection from image feature space to a semantic space through discriminative methods \cite{palatucci2009zero,akata2015label} or generative models \cite{long2017zero,wang2018zero}. Another popular way is to combine different modalities (such as images, texts, attributes, etc.) and learn a non-linear multi-modal embedding \cite{chaudhuri2020simplified, lin2020learning}, to help unseen class recognition. For example, Lin \textit{et. al.} \cite{lin2021zstgan} used a GAN-based model to learn a multi-modal consistent semantic representation, and the disentangled domain-invariant features are extracted for unsupervised zero-shot image-to-image translation.  

\textbf{Zero-shot Learning-based Cross-modality Data Translation}
The GAN-based zero-shot-learning cross-modality data translation models usually modify the latent representation of a pre-trained GAN model (a.k.a. GAN inversion \cite{ganinversion, SemanticStyleGAN, gan7, gan8}). It needs additional optimization in the testing step, which is computationally expensive and time-consuming.

Perturbation-diffusion-based cross-modality data translation allows Zero-shot learning for cross-modality data translation tasks \cite{adaptiveRealistic, DDPM, kawar2022denoising, songsde}. The perturbation-diffusion-based methods perform excellently when the numerical features of source and target domains are consistent \cite{sdeit}. Nevertheless, they may fail when the cross-domain appearance features are significantly distinct.


\section{Theroy}
\label{sec;preliminary}
\subsection{Cross-modality Data Translation}
Mathematically, the cross-modality data translation task between two modalities $\src \in \srcdomain$ and $\tar \in \tardomain$ can be formalized as:
\begin{equation}
    \res \in \tardomain:  \res = \mapping(G)
\end{equation}
, where $\mapping$ is an operator that maps the data $\src$ in the source domain $\srcdomain$ to the corresponding data $\res$, which is ideally the same as $\tar$ in the target domain $\tardomain$. Specifically, this generalized form has been broadly applied in image synthesis, stroke painting, image registration, segmentation, \textit{etc.} \cite{reviewI2I} In this paper, we will mainly focus on 2D image space. However, the proposed method can be extended to 1D or 3D signals without loss of generality.

In zero-shot cross-modality data translation, only samples in target domain $\tar \in \tardomain$ are available during the training phase. Given the target samples, the aim of Zero-shot-Learning is to learn $\mapping$ without seeing the $\src$ in training step. Since no source samples are accessible in the training phase, it is important to build and use auxiliary information for domain transfer \cite{wang2019survey}. 

\subsection{Mutual Information}
Mutual Information (MI) measures the dependence of two random variables $X,Y$:
\begin{equation}
\label{eq;MI}
        MI(X,Y) = \iint p(x, y) \log \frac{p(x,y)}{p(x)p(y)} dxdy
\end{equation}
MI is useful in cross-modality data processing tasks as the statistical features are assumed to be identical. It has been applied to tackle many unsupervised learning problems such as cross-modality data retrieval\cite{MICrossModal}, data representations \cite{hjelm2018learning,Zheng_2022_CVPR, pmlr-v162-guo22g}, domain adaptation \cite{MIDomainAdaptation}, and cross-modal clustering \cite{AAAIMI} \textit{etc.}. 

A particular case of MI is using MI for measuring a random variable itself: $MI(X,X)$, which is called \textit{Entropy}. 
\subsection{Score Matching and Its Denoising Equivalent}
Different from variational inference-based or likelihood-based training, which attempts to approximate the true probability distribution $\log p(x)$ of the data, the score-based models learn to represent the distribution $\log p(x)$ through its partial derivative $\score$ information (a.k.a score function). The maximization process between a learning model $\scorenet(x)$ and $\score$ needs to get an explicit form $q(x)$ of the accurate distribution $p(x)$, which usually remains unknown. 
Instead of seeking any form $q(x)$ of the distribution of the target dataset, the denoising score matching method \cite{score1} sidesteps the searching of $q(x)$ by directly estimating the score function through \cite{score1, score2}:
\begin{equation}
    \arg \min_\theta \mathbb{E}_{q_\sigma(x, \hat{x})}[\frac{1}{2}||s_\theta (x) - \dfrac{\partial \log q_\sigma (\hat{x}|x)}{\partial \hat{x}}||^2]
    \label{eq;loss}
\end{equation}
, where $\dfrac{\partial \log q_\sigma (\hat{x}|x)}{\partial \hat{x}}$ is the gradient between the clean data $x$ and a noise-polluted observation $\hat{x}$; as long as the noise is driven by a Gaussian kernel: $\dfrac{\partial \log q_\sigma (\hat{x}|x)}{\partial \hat{x}} = \frac{x - \hat{x}}{\sigma^2}$, the learning model $s(x)$ aims to learn the process of denoising.\cite{score1}  
\subsection{Score-based Generative Modeling with SDEs}
\label{sec;sdescore}
Song \textit{et. al.} \cite{songsde} generalized the above denoising score matching-based models into an SDE framework and unified the generative process by sampling with a reverse SDE \cite{ANDERSON1982313}, which is also a diffusion process. The neural network $\scorenet$ is trained in an implicit denoising form by treating the noise-adding steps as a diffusion process. The diffusion process driven by a standard normal distribution $\mathcal{N}(0,1)$ in a potential $U_t(x)$ can be modeled by an SDE:
\begin{equation}
    dx_t = U_t(x)dt + \sigma_t d\brwonian_t, ~~~~t \in [0, T]
    \label{eq;addingnoise}
\end{equation}
, where $\sigma_t \in [0, \infty)$ controls the magnitude of the input noise,  $\brwonian_t$ denotes a Wiener process, and $t \in [0, T]$ is the time start from 0 with infinitesimal incremental to $t$. We use a static potential $U_t(x)= 0 $ (a.k.a "Variance Exploding SDE (VE-SDE)" \cite{songsde}) for modality translation here, which models the magnitude of the data $x_0$ ( the target modality $\tar_0$). Adding noise with Eq. \ref{eq;addingnoise} is a similar step when using a Gaussian transition kernel to perform score matching in Eq. \ref{eq;loss}. The training target of a diffusion process defined by the SDE \ref{eq;addingnoise} shares a similar training target as Eq. \ref{eq;loss} yet with an expectation of time $t$ sampled uniformly between [0, T]: $\mathbb{E}_{t\sim \mathcal{U}(0, T)}$; also the whole process becomes a multi-step conditioning $q(x_t|x_0)$ instead of a single step $q(\hat{x}|x)$,
\begin{multline}
    \arg \min_\theta \mathbb{E}_{t\sim \mathcal{U}(0,T)}
    \{ \mathbb{E}_{x_0 \sim p_0(x)}  \\ 
    \mathbb{E}_{x_t \sim q_{\sigma_t}(x_t, x_0)}[\frac{1}{2}||s_\theta (x) - \dfrac{\partial\log q_{\sigma_t} (x_t|x_0)}{\partial x_t}||^2] \}
    \label{eq;losssde}
\end{multline}
\begin{figure}
    \centering
    \includegraphics[width=0.85\textwidth]{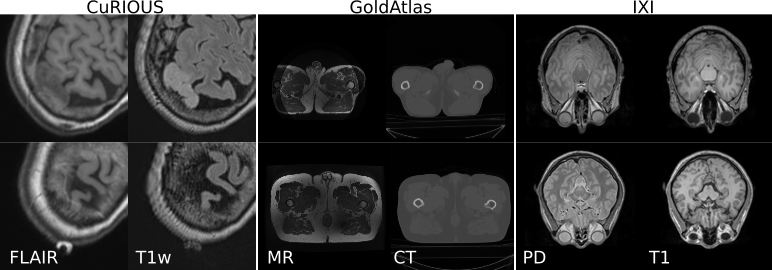}
    \caption{Example data of different modalities for the three different datasets; from left to right are: \datab~of FLAIR imaging, \datab~of T1-weighted (T1-w) imaging, \dataa~of MR imaging,  \dataa~of CT imaging, \datac~of PD-weighted imaging, \datac~of T1-w imaging.}
    \label{fig;dataset}
\end{figure}
As long as the matched score model $\scorenet \sim \score$ learns the distribution score $\score$, we can employ a reverse SDE \cite{songsde, ANDERSON1982313, sdeit} to sample datapoints by inferring a backward in time $t: T \rightarrow 0$ dynamic process:
\begin{align}
\label{eq;backward}
    dx_t &= - \frac{d\sigma_t^2}{dt} [\score]_t dt + \sqrt{\frac{d\sigma_t^2}{dt}} d\brwonian_t,\\
    &= - \frac{d\sigma_t^2}{dt} \scorenet(x_t, t) dt + \sqrt{\frac{d\sigma_t^2}{dt}} d\brwonian_t, ~~~~t_{T \rightarrow 0} \in [0, T]
\end{align}
, where $\brwonian_t$ is a Wiener process with infinitesimal negative incremental of time $dt$. 
\begin{figure*}
    \centering
    \includegraphics[width=\textwidth]{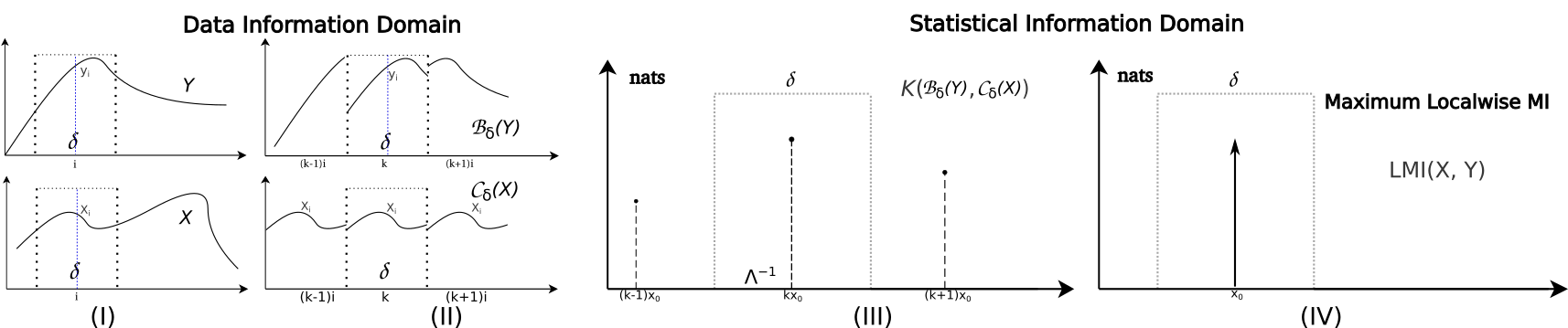}
    \caption{Illustration of an example of the proposed operator functioning for 1D functions $X$ and $Y$ in the neighborhood $\delta$ of point $x_i$. (I) $\delta$-neighborhood of the two functions (upper $Y$, lower $X$) at point $y_i$ and $x_i$. (II) functions are processed by the proposed operators defined in Def. \ref{def;linear} and Def. \ref{def;nearst}. (III) compute the Mutual Information between the paired segments in step (II) using kernel density estimation. (IV) select the maximum value of MI in (III) as $LMI_{\delta}(x_i, y_i)$ at point $x_i$ between $X$ and $Y$.}
    \label{fig;operator}
\end{figure*}
\section{Methods}
\subsection{Diffusion for Cross-modality Data Translation}
We can solve the cross-modality data translation problem by adapting the target data $\tar$ generation task into the framework of score-matching and then using a perturbed source domain $\src$ to guide (conditioning) the iterative diffusion process \cite{sdeit, GANDIFF, diffi2i1, ozdenizci2022, EnergyGuided, kawar2022denoising, nonuni}. Ideally, we want the generated data $\res$ to follow the semantic meaning of the guided data $\src$ and share the features as the ground truth data point $\tar$ in the target domain, balancing  \textbf{realism} and \textbf{faithfulness}.\cite{sdeit, adaptiveRealistic}
\subsubsection{Cross-modality Data Translation Fidelity}
\label{sec;fidelity}
\begin{defi}
\label{def;real}
The generated image $\res$ shows \textbf{realism} means it is well translated into the target domain $\tardomain$: $\res \in \tardomain$.
\end{defi}
\begin{defi}
\label{def;faith}
The generated image $\res$ remains \textbf{faithfulness} means it is faithfully translated from the guided data: $\res \sim \src \in \srcdomain$, where $\sim $ is a similarity measure.
\end{defi}
The relationship between translation realism and faithfulness is related to the domain correlations $\tardomain$ and $\srcdomain$.
We say a translation achieves high \textbf{fidelity} if and only if the translation achieves a balance point between the \textbf{realism} and \textbf{faithfulness}.
However, when the difference in the appearance feature set between the source domain $\src$ and target domain $\tar$ becomes too large, a balance point between the two sides may neither be easy to capture nor exist for a satisfying translation. This leads current distribution perturbing-based methods (e.g., SDEdit, \textit{etc.} \cite{sdeit, score1, score2}) unsuitable for unsupervised cross-modality data translation when the numerical features between the two domains have huge differences.

\subsection{Mutual Information Guidance in Diffusion Generation}
In the Zero-shot-Learning-based translation task, we do not have access to the data in the source domain during the training process. Nevertheless, the local statistical features between the source and target modalities are assumed to be identical. MI maximization has been proven an effective method to empower neural networks to learn non-linear representations \cite{hjelm2018learning}. To capture those shared representations and use the extracted information for generation guidance, we propose using MI to measure the local statistical representations in the iterative denoising process. 

\subsubsection{Local-wise MI}
To obtain the semantic information in the data for guidance, we need to convert the original data to statistical representation, as MI is a statistical measure.
Given a datapoint $X$, for point $x_i \in X$ at position $i$, the local-wise statistical information at $i$ can be captured through the probability density function (PDF) $p_{\delta_{x_i}}(\cdot)$ of the neighborhood area $\delta_{x_i}$ of $x_i$. 

Without loss of generality, for the other points $x_j$ inside the neighborhood area $\delta_{x_i}$ of $x_i$, we can obtain the local-wise statistical information through the PDF $p_{\delta_{x_j}}(\cdot); j \in \delta_i$.
\begin{defi}
The local-wise MI ($LMI$) from data point $X$ to data point $Y$ at point $x_i$ is defined through:
\begin{multline}
\label{eq;LMI}
        LMI_\delta(x_i,y_j) = \\
        \sup \iint p_\delta(x, y) \log \frac{p_\delta(x, y)}{p_{\delta_{x_i}}(x)p_{\delta_{y_j}}(y)} dx dy, \forall y_j \in \delta_{x_i}    
\end{multline}
\end{defi}
In the forward steps (training), we can use Eq. \ref{eq;LMI} as a reference signal to condition each diffusion step; this is achieved by computing the $LMI(X_0, X_t), t \in [0, T]$ during the training process of the scoring neural network.
\begin{theo}
\label{theo;optmum}
The upper bound of the $LMI$ from data point $X$ to datapoint $Y$ at location $i$ is: $LMI_\delta(x_i,y_i) \leq LMI_\delta(x_i,x_i)$
, which is the optimum informative match between $X$ and $Y$ at point $x_i$.
\end{theo}

The Thm. \ref{theo;optmum} indicates that the $LMI$ achieves maximum (statistical similarity) when: $p_\delta(X) = p_\delta(Y), \forall \delta \in X$. Thus the $LMI$ always achieves maximum at the same position between $X_0$ and $X_t$ in the training step; Yet, when $p_\delta(X) \neq p_\delta(Y)$, the $LMI$ achieves the local maximum at position $j$, which is located in the neighborhood $\delta_{y_i}$ of $y_i$.

We have defined the $LMI$ for conditioning the iterative diffusion process. However, the computational cost of $LMI$ is very high as a statistical measure. The iterative training of the score model requires many steps of discrete time conditioning. It is unrealistic to use a sliding window or patch-wise looping to compute the $LMI$ between each time steps. To overcome this bottleneck of applying $LMI$ in score matching models, we developed an efficient method to compute the $LMI$ in both of the perturbing and denoising steps.


\subsubsection{Differentiable local-wise MI Layer}
\begin{defi}
Let $X(i)$ be a function defined in $\mathbb{R}^n$, then $\mathcal{C}_{\delta}(X) \in \mathbb{R}^n$ is a \textbf{periodic extension} of $X$ in neighborhood $K\cdot\delta, K \in \mathbb{Z}$:
\begin{multline}
\mathcal{C}_{\delta}(X):=     \begin{cases}
        X & for~i \in \delta \\
        is~periodic~of~ \delta & for ~i \in K \cdot \delta\\
        0 & otherwise\\
    \end{cases}
\end{multline}
\label{def;nearst}
\end{defi}

\begin{defi}
Let $Y(i)$ be a function defined in $\mathbb{R}^n$, then $\mathcal{B}_{\delta}(Y) \in \mathbb{R}^n$ is a $k-$steps ($k \in \mathbb{Z}^n$) \textbf{sliding extension} of $Y$ in neighborhood $k \cdot \delta$:
\begin{multline}
\mathcal{B}_{\delta}(Y):=J_\tau Y(i-\tau)\\ 
s.t.~~~~ \mathbf{J}_{\tau} = \begin{cases}
        1 &  \tau - \frac{\delta}{2} \leq i < \tau + \frac{\delta}{2}\\
        0 & else 
    \end{cases}
\end{multline}
\label{def;linear}
for $\tau: 0 \rightarrow \delta$ with incremental $\Delta\tau = \delta/k$.
\end{defi}


Definition (\ref{def;linear}) and (\ref{def;nearst}) can be understood as a 'segment-wise' linear interpolation and nearest-neighbor interpolation \cite{imagewarp} of functions $Y$ and $X$ in the neighborhood $\delta$. A demonstration of the operator applied for 1D functions defined in ( Def. \ref{def;nearst}) and (Def. \ref{def;linear}) is shown in Fig. \ref{fig;operator} (II), for which the two functions $X$ and $Y$ (Fig. \ref{fig;operator} (I)) are processed by the operators $\mathcal{C}$ and $\mathcal{B}$ respectively. 
\begin{prop}
\label{prop1}
With two given functions $\tar \in \tardomain$ and $\src \in \tardomain$ 
in $\mathbb{R}^2$, the $LMI$ (Def. \ref{eq;LMI}) between $\src$ and $\tar$ can be computed through the following operator $\mathcal{S}$:
\begin{multline}
    LMI(G,F) = \max_\delta(\mathcal{K}(\mathcal{B}_{\delta}({F}),\mathcal{C}_{\delta}(G))\log\frac{\mathcal{K}(\mathcal{B}_{\delta}({F}),\mathcal{C}_{\delta}(G))}{\mathcal{K}\mathcal{B}_{\delta}({F}) \mathcal{K}\mathcal{C}_{\delta}(G))}) \\
     \delta \in \src, \tar
\end{multline}
where $\mathcal{K}$ is a kernel density estimator, which approximates the PDF.
\end{prop}
The Prop. \ref{prop1} computes the maximum $LMI$ in both of the training and testing steps, which are $LMI_{\delta}(F,F)$ and $LMI_{\delta}(G,F)$ respectively. The operator given in Prop. \ref{prop1} can be accelerated by memory copying and parallel reduction in GPGPU \cite{cudac}.
\begin{algorithm}
\SetAlgoLined
\SetKwInput{KwData}{Inputs}
\SetKwInput{KwResult}{Output}
\KwData{$\src$, $\scorenet$, time $t_{T \rightarrow 0} \in [0, T]$, step size $dt$}
\KwResult{Generated high fidelity data $\res$}
 Initialize all parameters, variables\;
\While{t < T}{
    $z \sim \mathcal{N}(0, \textbf{1})$\; 
    $t \leftarrow t + dt$\; 
    $\epsilon = \sigma_t^2 - \sigma^2_{t-dt}$\; 
    $\res_{t+1} \leftarrow \res_{t} + \epsilon \scorenet(\res_{t}, LMI(\src; \src, \res_{t}), t) + \sqrt{\epsilon} z$ \tcp*{Euler discretization step, guided by $\src$}
}
\caption{Mutual Information Guided Diffusion}
\label{alg;solver}
\end{algorithm}
\subsection{Embedding the Conditioner into SDE}
In the training step (see left sub-figure of Fig. \ref{fig:overview}), we can condition the noise perturbing process by embedding the defined $LMI$ between a datapoint $\tar$ and its disturbed data $\tar_t$ in to the time-dependent-scoring model $\scorenet$, which is guided by $\src$:
\begin{multline}
    \arg \min_\theta \mathbb{E}_{t\sim \mathcal{U}(0,T)}
    \{ \mathbb{E}_{x_0 \sim p_0(x)} 
    \mathbb{E}_{x_t \sim q_{\sigma_t}(x_t, x_0)}\\
    [\frac{1}{2}| \scorenet(\res_t, LMI(\tar; \tar, \tar_t), t) - \dfrac{\partial \log q_{\sigma_t} (x_t|x_0)}{\partial x_t}|^2] \}
    \label{eq;losssdeLMI}
\end{multline}

In the sampling step (see right sub-figure of Fig. \ref{fig:overview}), recall that the generative process is an iterative solution of the reverse SDE \ref{eq;backward}. We can inject the proposed conditioning procedure into the SDE \ref{eq;backward}:
\begin{multline}
\label{eq;embedded}
    d\res_t =- \frac{d\sigma_t^2}{dt} \scorenet(\res_t, LMI(\src; \src, \res_t), t) dt + \\
    \sqrt{\frac{d\sigma_t^2}{dt}} d\brwonian_t, ~~~~t_{T \rightarrow 0} \in [0, T]
\end{multline}
and use an naive Euler-Maruyama numerical solver (see Algorithm \ref{alg;solver}) to solve the SDE.

\begin{table}
\centering
\caption{Quantitative evaluation of the MIDiffusion in comparison with three baselines on three different cross-modality data translation tasks. The best-performing values on the same task being compared are bolded in black italics for highlighting. sup=supervised; unsup=unsupervised}
\label{tab;results}
\resizebox{\textwidth}{!}{\begin{tabular}{ccccccccc}
\textbf{Datasets}                                            & \textbf{Methods}                                                                                                                                                                                                                                                                                                                                                               & \textbf{Modalities}                                                                                                                                   & \begin{tabular}[c]{@{}c@{}}\textbf{\textbf{SSIM}}\\\textbf{\textbf{Tar$\uparrow$}}\end{tabular} & \begin{tabular}[c]{@{}c@{}}\textbf{SSIM}\\\textbf{Src$\uparrow$}\end{tabular} & \textbf{\textbf{MSE$\downarrow$}}                    & \textbf{\textbf{MI$\uparrow$}}                    & \textbf{\textbf{PSNR$\uparrow$}}                   & \textbf{\textbf{FID$\downarrow$}}  \\ 
\hline\hline
\multirow{8}{*}{\textbf{\textbf{GoldAtlas}}}                 & \multirow{2}{*}{\begin{tabular}[c]{@{}c@{}}\textbf{\textbf{CycleGAN}}\cite{CycleGAN2017}\\\textbf{\textbf{(\textbf{\textbf{\textbf{\textbf{\textbf{\textbf{sup,~}}}}}}few-shot 2\%)}}\end{tabular}}                                                                                                                                                            & \textbf{\textbf{CT→MR}}                                                                                                                               & 0.04                                                                                            & 0.03                                                                          & 614.02                                               & \textbf{\textbf{\textit{1.16}}}                   & 20.53                                              & \textit{\textbf{202.43}}           \\
                                                             &                                                                                                                                                                                                                                                                                                                                                                                & \textbf{\textbf{MR→CT}}                                                                                                                               & 0.03                                                                                            & 0.02                                                                          & 819.59                                               & 1.13                                              & 19.08                                              & 281.35                             \\
                                                             & \multirow{2}{*}{\begin{tabular}[c]{@{}c@{}}\textbf{\textbf{\textbf{\textbf{StyleGAN~}}}}~\cite{styleGAN2}\\\textbf{\textbf{\textbf{\textbf{(unsup,~inversion)}}}}\end{tabular}}                                                                                                                                                                               & \textbf{\textbf{\textbf{\textbf{CT→MR}}}}                                                                                                             & \textbf{\textbf{\textit{0.13}}}                                                                 & 0.04                                                                          & 788.76                                               & 1.09                                              & 20.09                                              & 213.47                             \\
                                                             &                                                                                                                                                                                                                                                                                                                                                                                & \textbf{\textbf{MR→CT}}                                                                                                                               & 0.08                                                                                            & 0.07                                                                          & 570.91                                               & 1.12                                              & 21.17                                              & \textit{\textbf{170.83}}           \\
                                                             & \multirow{2}{*}{\begin{tabular}[c]{@{}c@{}}\textbf{\textbf{\textbf{\textbf{\textbf{\textbf{\textbf{\textbf{SDEdit}}}}}}}}~\cite{sdeit}\textbf{\textbf{\textbf{\textbf{\textbf{\textbf{\textbf{\textbf{}}}}}}}}\\\textbf{\textbf{\textbf{\textbf{\textbf{\textbf{\textbf{\textbf{(unsup)}}}}}}}}\end{tabular}}                                                 & \textbf{\textbf{\textbf{\textbf{\textbf{\textbf{\textbf{\textbf{CT→MR}}}}}}}}                                                                         & 0.003                                                                                           & 0.01                                                                          & 766.40                                               & 1.11                                              & 19.50                                              & 237.27                             \\
                                                             &                                                                                                                                                                                                                                                                                                                                                                                & \textbf{\textbf{\textbf{\textbf{MR→CT}}}}                                                                                                             & 0.01                                                                                            & 0.04                                                                          & 996.71                                               & 1.10                                              & 18.58                                              & 223.44                             \\
                                                             & \multirow{2}{*}{\begin{tabular}[c]{@{}c@{}}\textbf{\textbf{\textbf{\textbf{MIDiffusion}}}}\\\textbf{\textbf{\textbf{\textbf{\textbf{\textbf{\textbf{\textbf{(unsup)}}}}}}}}\end{tabular}}                                                                                                                                                                                      & \textbf{\textbf{\textbf{\textbf{\textbf{\textbf{\textbf{\textbf{\textbf{\textbf{\textbf{\textbf{\textbf{\textbf{\textbf{\textbf{CT→MR}}}}}}}}}}}}}}}} & 0.06                                                                                            & \textit{\textbf{0.11}}                                                        & \textbf{\textbf{\textit{523.18}}}                    & 1.08                                              & \textbf{\textbf{\textit{21.66}}}                   & 245.82                             \\
                                                             &                                                                                                                                                                                                                                                                                                                                                                                & \textbf{\textbf{\textbf{\textbf{\textbf{\textbf{\textbf{\textbf{MR→CT}}}}}}}}                                                                         & \textbf{\textbf{\textit{0.12}}}                                                                 & \textit{\textbf{0.08}}                                                        & \textbf{\textbf{\textit{392.35}}}                    & \textbf{\textbf{\textit{1.17}}}                   & \textbf{\textbf{\textit{23.03}}}                   & {194.35}                     \\ 
\hline
\multirow{8}{*}{\textbf{\textbf{\textbf{\textbf{CuRIOUS}}}}} & \multirow{2}{*}{\begin{tabular}[c]{@{}c@{}}\textbf{\textbf{\textbf{\textbf{CycleGAN }}}}\\\textbf{\textbf{\textbf{\textbf{(\textbf{\textbf{\textbf{\textbf{sup,~}}}}few-shot \textasciitilde{}6\%)}}}}\end{tabular}}                                                                                                                                                            & \textbf{\textbf{T1→FLAIR}}                                                                                                                            & -0.006                                                                                          & 0.81                                                                          & 1747.13                                              & \textbf{\textit{1.08}}                            & 16.04                                              & 186.59                             \\
                                                             &                                                                                                                                                                                                                                                                                                                                                                                & \textbf{\textbf{\textbf{\textbf{FLAIR}}→T1}}                                                                                                          & 0.005                                                                                           & 0.02                                                                          & 3145.05                                              & 1.05                                              & 13.82                                              & 331.89                             \\
                                                             & \multirow{2}{*}{\begin{tabular}[c]{@{}c@{}}\textbf{\textbf{\textbf{\textbf{\textbf{\textbf{\textbf{\textbf{StyleGAN }}}}}}}}\\\textbf{\textbf{\textbf{\textbf{\textbf{\textbf{\textbf{\textbf{(unsup,~inversion)}}}}}}}}\end{tabular}}                                                                                                                                         & \textbf{\textbf{T1→\textbf{\textbf{FLAIR}}}}                                                                                                          & 0.003                                                                                           & 0.12                                                                          & 1880.62                                              & 1.04                                              & 15.83                                              & 261.47                             \\
                                                             &                                                                                                                                                                                                                                                                                                                                                                                & \textbf{\textbf{\textbf{\textbf{FLAIR}}→T1}}                                                                                                          & -0.003                                                                                          & 0.19                                                                          & 1570.83                                              & 1.05                                              & 16.62                                              & 229.73                             \\
                                                             & \multirow{2}{*}{\begin{tabular}[c]{@{}c@{}}\textbf{\textbf{\textbf{\textbf{\textbf{\textbf{\textbf{\textbf{\textbf{\textbf{\textbf{\textbf{\textbf{\textbf{\textbf{\textbf{SDEdit }}}}}}}}}}}}}}}}\\\textbf{\textbf{\textbf{\textbf{\textbf{\textbf{\textbf{\textbf{\textbf{\textbf{\textbf{\textbf{\textbf{\textbf{\textbf{\textbf{(unsup)}}}}}}}}}}}}}}}}\end{tabular}}      & \textbf{\textbf{T1→\textbf{\textbf{FLAIR}}}}                                                                                                          & 0.011                                                                                           & 0.01                                                                          & 1558.22                                              & 1.04                                              & 16.42                                              & \textit{\textbf{131.70}}           \\
                                                             &                                                                                                                                                                                                                                                                                                                                                                                & \textbf{\textbf{\textbf{\textbf{FLAIR}}→T1}}                                                                                                          & 0.005                                                                                           & 0.01                                                                          & 2165.42                                              & 1.03                                              & 15.14                                              & \textbf{\textit{141.89}}           \\
                                                             & \multirow{2}{*}{\begin{tabular}[c]{@{}c@{}}\textbf{\textbf{\textbf{\textbf{\textbf{\textbf{\textbf{\textbf{MIDiffusion~\textbf{\textbf{\textbf{\textbf{\textbf{\textbf{\textbf{\textbf{}}}}}}}}}}}}}}}}\\\textbf{\textbf{\textbf{\textbf{\textbf{\textbf{\textbf{\textbf{\textbf{\textbf{\textbf{\textbf{\textbf{\textbf{\textbf{\textbf{(unsup)}}}}}}}}}}}}}}}}\end{tabular}} & \textbf{\textbf{T1→\textbf{\textbf{FLAIR}}}}                                                                                                          & \textit{\textbf{0.07}}                                                                          & -0.08                                                                         & \textit{\textbf{1226.40}}                            & \textit{\textbf{1.08}}                            & \textit{\textbf{17.65}}                            & {146.77}                     \\
                                                             &                                                                                                                                                                                                                                                                                                                                                                                & \textbf{\textbf{\textbf{\textbf{FLAIR→T1}}}}                                                                                                          & \textbf{\textbf{\textbf{\textbf{\textit{0.15}}}}}                                               & \textit{\textbf{0.23}}                                                        & \textbf{\textbf{\textbf{\textbf{\textit{1175.11}}}}} & \textbf{\textbf{\textbf{\textbf{\textit{1.08}}}}} & \textbf{\textbf{\textbf{\textbf{\textit{18.02}}}}} & {157.98}                     \\ 
\hline
\multirow{8}{*}{\textbf{\textbf{\textbf{\textbf{IXI}}}}}     & \multirow{2}{*}{\begin{tabular}[c]{@{}c@{}}\textbf{\textbf{\textbf{\textbf{CycleGAN }}}}\\\textbf{\textbf{\textbf{\textbf{(sup, few-shot 11\%)}}}}\end{tabular}}                                                                                                                                                                                                                & \textbf{\textbf{PD\textbf{\textbf{→}}T1}}                                                                                                             & \textbf{\textit{0.12}}                                                                          & 0.14                                                                          & \textbf{\textit{1154.19}}                            & \textbf{\textit{1.17}}                            & \textbf{\textit{17.65}}                            & 141.95                             \\
                                                             &                                                                                                                                                                                                                                                                                                                                                                                & \textbf{\textbf{T1→PD}}                                                                                                                               & 0.16                                                                                            & 0.16                                                                          & \textbf{\textit{876.99}}                             & \textit{\textbf{1.19}}                            & \textbf{\textit{18.86}}                            & 113.67                             \\
                                                             & \multirow{2}{*}{\begin{tabular}[c]{@{}c@{}}\textbf{\textbf{\textbf{\textbf{\textbf{\textbf{\textbf{\textbf{StyleGAN }}}}}}}}\\\textbf{\textbf{\textbf{\textbf{\textbf{\textbf{\textbf{\textbf{(unsup,~inversion)}}}}}}}}~\end{tabular}}                                                                                                                                        & \textbf{\textbf{PD-T1}}                                                                                                                               & 0.02                                                                                            & 0.06                                                                          & 6609.13                                              & 1.08                                              & 10.17                                              & 266.52                             \\
                                                             &                                                                                                                                                                                                                                                                                                                                                                                & \textbf{\textbf{T1→PD}}                                                                                                                               & \textbf{\textit{0.21}}                                                                          & \textit{\textbf{0.37}}                                                        & 2319.78                                              & \textit{1.14}                                     & 14.65                                              & 199.12                             \\
                                                             & \multirow{2}{*}{\begin{tabular}[c]{@{}c@{}}\textbf{\textbf{\textbf{\textbf{\textbf{\textbf{\textbf{\textbf{\textbf{\textbf{\textbf{\textbf{\textbf{\textbf{\textbf{\textbf{SDEdit }}}}}}}}}}}}}}}}\\\textbf{\textbf{\textbf{\textbf{\textbf{\textbf{\textbf{\textbf{\textbf{\textbf{\textbf{\textbf{\textbf{\textbf{\textbf{\textbf{(unsup)}}}}}}}}}}}}}}}}\end{tabular}}      & \textbf{\textbf{PD-T1}}                                                                                                                               & 0.09                                                                                            & 0.06                                                                          & 1619.14                                              & 1.15                                              & 16.19                                              & \textbf{\textit{68.6}}             \\
                                                             &                                                                                                                                                                                                                                                                                                                                                                                & \textbf{\textbf{T1→PD}}                                                                                                                               & 0.10                                                                                            & 0.06                                                                          & 1753.82                                              & 1.16                                              & 15.95                                              & \textit{\textbf{80.81}}            \\
                                                             & \multirow{2}{*}{\begin{tabular}[c]{@{}c@{}}\textbf{\textbf{\textbf{\textbf{\textbf{\textbf{\textbf{\textbf{MIDiffusion}}}}}}}}\\\textbf{\textbf{\textbf{\textbf{\textbf{\textbf{\textbf{\textbf{~\textbf{\textbf{\textbf{\textbf{\textbf{\textbf{\textbf{\textbf{(unsup)}}}}}}}}}}}}}}}}\end{tabular}}                                                                         & \textbf{\textbf{PD-T1}}                                                                                                                               & \textit{0.11}                                                                                   & \textit{\textbf{0.19}}                                                        & \textit{1652.81}                                     & \textbf{\textit{1.17}}                            & 16.35                                              & {129.12}                     \\
                                                             &                                                                                                                                                                                                                                                                                                                                                                                & \textbf{\textbf{\textbf{\textbf{T1→PD}}}}                                                                                                             & 0.18                                                                                            & \textit{0.26}                                                                 & \textit{1301.91}                                     & 1.13                                              & 17.13                                              & {132.46}                     \\
\hline\hline
\end{tabular}}
\end{table}
\section{Experiments and Evaluation}
We show the following features of the proposed method in this section:
(1) The proposed zero-shot unsupervised learning method outperforms GAN-based cross-modality data translation models in both the zero-shot and few-shot supervised tracks. (2) Our method achieves high cross-modality data translation fidelity compared with the state-of-the-art diffusion method. (3) The proposed $LMI$-guided diffusion model enables cross-modality data translation semantic consistency in the practical application.
\subsection{Evaluation Datasets}
We introduce three public datasets that cover different cross-modality data translation tasks. Fig. \ref{fig;dataset} shows six samples collected from three different datasets:
(1) The \dataa~includes CT and T1-weighted (T1w) and T2-weighed (T2w) magnetic resonance imaging (MRI) data from 19 male patients. The pelvic area was the subject of the imaging. All the CT images had been deformably registered to the corresponding MRI \cite{nyholm_tufve_2017_583096}. The datasets were collected from three different sites \cite{Nyholm2018}. We resampled all the datasets using the SimpleITK package to a uniform voxel size  $(0.875mm \times 0.875 mm \times 3mm)$, and selected slices  every $15mm$ interval to build the training set ($993$ CT-MR slice pairs from 15 patients) and the testing set ($227$ CT-MR slice pairs from 4 patients).

(2) The publicly available \datab~consists of 22  subjects with low-grade glioma \cite{8795512}. The original dataset includes T1w and FLAIR MRI scan pairs, as well as unregistered ultrasound scans. All the images were collected during routine clinical exams. The original scans had been resampled to  $256 \times 256 \times 288$ voxels at an isotropic voxel size of $0.5mm^3$ \cite{curious1}. We chose the T1w and FLAIR MRI scan pairs to build our cross-modality image translation task. All the volume pairs were resampled to $128 \times 128 \times 288$ voxels with a spacing size of $1.0mm \times 1.0mm \times 0.5mm$ by the SimpleITK package. As a result, $1168$ FLAIR-T1w slice pairs were picked in every $5mm$ from the original voxels to generate the training ($952$ pairs from 17 subjects) and testing ($216$ pairs from 5 subjects) datasets.

(3) \datac\footnote{https://brain-development.org/ixi-dataset/}~includes pre-aligned 600 images from normal subjects. The full dataset includes five different modalities: T1w, T2w, PD-weighted, magnetic resonance angiograms (MRA), and Diffusion-weighted MRI. In our experiment, we perform PD-T1w modal translation tasks. A training set of 300 slices (sampled from 100 subjects) and a testing set of 75 slices (sampled from 25 subjects) were generated using a subset of the \datac~.
\subsection{Baselines}
We select two representative GAN-based image translation and synthesis approaches and one state-of-the-art diffusion-based method as baselines for comparison. We first compare our zero-shot-Learning-based method with a few-shot learning-based CycleGAN translation model \cite{CycleGAN2017}. The CycleGAN in this experiment will be allowed to see both the full target domain dataset and a small group (about $2\%$ for \dataa, $6\%$ for \datab and $11\%$ for \datac) of the source domain dataset. However, our proposed method will  see only the data in the target domain.

The second baseline method is a GAN inversion-based approach. A StyleGAN2-ADA \cite{styleGAN2} is allowed to see the target domain training data. The out-domain guided generation is performed through 5000 steps of optimization of inversion in the latent space of the trained StyleGAN2-ADA \cite{ganinversion}. An extra encoder network is introduced in the generation step for inversion.

The third baseline model is SDEdit \cite{sdeit}, which works for an ablation study. The SDEdit is also a diffusion model-based translation method but uses a distribution perturbation guidance. 
Whenever possible, we use a default training setup for all the baselines provided in the papers. The experiments were implemented based on publicly available open-source code in the same experiment environment. 
\begin{figure*}
    \centering
    \includegraphics[width=\textwidth]{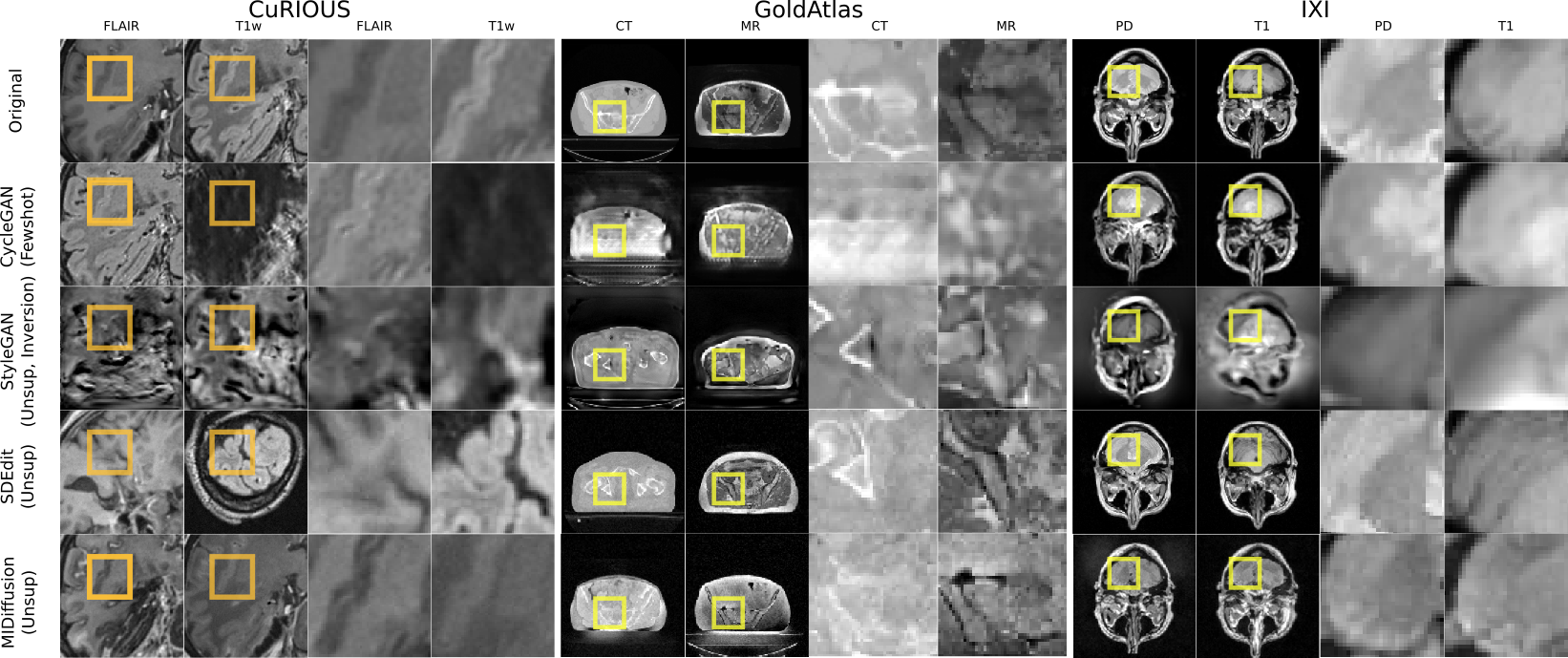}
    \caption{Qualitative performance of different methods on CuRIOUS, GoldAtlas, and IXI datasets. The top row shows the original images. The second to the last rows show: CycleGAN (few-shot sup), StyleGAN(unsup zero-shot with GAN inversion), SDEdit(zero-shot unsup learning with SDE-based perturbing guidance), and the proposed MIDiffusion (unsup zero-shot with $LMI$ guidance).}
    \label{fig;results}
\end{figure*}
\subsection{Quantitative Performance}
We evaluate the different methods based on five metrics, including three image quality measures: SSIM (structural similarity index measure), PSNR (peak signal-to-noise ratio), MSE (mean square error), and two statistical similarity measures: FID (Fr\'{e}chet Inception Distance) \cite{FID} and MI (mutual information). To study translation faithfulness (Def.~\ref{def;faith}) of each method, we report the SSIM between the translation results $\res$ and the guidance data $\src$ (SSIM-Src) and the target data $\tar$ (SSIM-Tar). 
The FID and MI are used to evaluate the translation similarity (Def.~\ref{def;real}) statistically between $\res$ and $\tar$ for different methods. The FID evaluates the generation similarity in the feature level between the $\res$ and $\tar$. The MI assesses the statistical similarity between the paired $\res$ and $\tar$ images. We show the performance based on those metrics of the four methods in Table. \ref{tab;results}.
\subsubsection{Comparison with few-shot training} 
We compare our method with the supervised few-shot trained CycleGAN (source domain $\tar \in \tardomain$ is visible) on each dataset. We notice that the size of the source datasets for the few-shot training differs between different datasets.
Regarding the translation errors (measured by SSIM, MSE, and PSNR), the proposed zero-shot trained MIDiffusion model outperforms the few-shot trained CycleGAN model on both the GoldAtlas (2\% of training data from source, 100\% from target) and CuRIOUS  (6\% of training data from source,, 100\% from target) datasets. However, this advantage of the MIDiffusion vanishes on the IXI dataset for the MSE and PSNR metrics. This can be attributed to the 11\% training data from the source domain being sufficient for the CycleGAN to surpass the performance of the zero-shot-learning models.
\subsubsection{Comparison with zero-shot training} 
As shown in Table \ref{tab;results}, the proposed method performs better than the other one-shot training-based methods in terms of the SSIM-Src, SSIM-Tar, and MSE, which means the generated data of MIDiffusion achieves higher semantic faithfulness with regard to both the source and target domain.
Our model also outperforms the other methods in terms of the PSNR and MI. This implies our method shows translation realism \ref{def;real} with respect to the target data in pairwise comparison. In terms of similarity between the generated dataset and the target dataset, we see that the SDEdit has an overall lower FID score. Yet, in considering our generation target of fidelity translation, the SDEdit fails in faithfulness translation. Thus the translation results of SDEdit performs worse.
In contrast, the proposed method achieves good realism (lowest FID score apart from SDEdit) while keeping the semantic meaning from the guidance to the target (higher SSIM, lower MSE). The cross-modality data translation method with the highest fidelity is therefore the MIDiffusion. 
\subsection{Qualitative Performance}
Fig. \ref{fig;results} shows translated results from different methods on the test dataset as well as the groundtruth (top rows). The first two columns in each dataset group represent different modalities, and the last two columns correspond to their zoomed-in details. Overall, the proposed method achieves the best translation fidelity among the compared methods. The anatomical structures are more faithfully represented in the MIDiffusion translation results compared to the results from the baseline models. The SDEdit and StyleGAN fail to translate the source images with identical features. In addition, the few-shot trained CycleGAN clearly underperforms the zero-shot trained MIDiffusion model when trained with insufficient source data on the CuRIOUS (2\%) and GoldAtlas (6\%) dataset. This observation is consistent with the quantitative results shown in Table \ref{tab;results}.
Overall, the MIDiffusion method outperforms the baseline methods in both anatomical consistency and appearance similarity.  
\section{Conclusion}
This paper presents a novel local-wise mutual information-guided diffusion model named MIDiffusion for cross-modality data translation. Different from current cycle-consistency training, the MIDiffusion does not require seeing the source dataset for training. Unlike GAN inversion methods that require iterative optimization during the generation, the MIDiffusion does not need online optimization in the test steps.
Our method introduces a new conditioner that achieves high-fidelity generation but does not need any extra training  on the main diffusion flow. This feature is different from supervised guidance \cite{dhariwal2021diffusion}, which needs to train an extra model to pose conditional guidance on the diffusion flow. The proposed diffusion model is more robust than another diffusion model (SDEdit) in terms of translation faithfulness, thanks to the $LMI$ guidance signal of MIDiffusion.

\textbf{Limitations} However, the MIDiffusion requires hundreds of times iterative solutions of SDE, which takes dozens of seconds for translating a single image on a GPU. Effectively reducing the number of sampling steps without compromising the translation fidelity would be meaningful for future work.

\bibliographystyle{abbrv}
\bibliography{main}

\begin{thebibliography}{10}

\bibitem{gan8}
R.~Abdal, Y.~Qin, and P.~Wonka.
\newblock Image2stylegan++: How to edit the embedded images?
\newblock In {\em Proceedings of the IEEE/CVF Conference on Computer Vision and
  Pattern Recognition (CVPR)}, June 2020.

\bibitem{akata2015label}
Z.~Akata, F.~Perronnin, Z.~Harchaoui, and C.~Schmid.
\newblock Label-embedding for image classification.
\newblock {\em IEEE transactions on pattern analysis and machine intelligence},
  38(7):1425--1438, 2015.

\bibitem{ANDERSON1982313}
B.~D. Anderson.
\newblock Reverse-time diffusion equation models.
\newblock {\em Stochastic Processes and their Applications}, 12(3):313--326,
  1982.

\bibitem{cmtgan7}
M.~Arar, Y.~Ginger, D.~Danon, A.~H. Bermano, and D.~Cohen-Or.
\newblock Unsupervised multi-modal image registration via geometry preserving
  image-to-image translation.
\newblock In {\em Proceedings of the IEEE/CVF Conference on Computer Vision and
  Pattern Recognition (CVPR)}, June 2020.

\bibitem{gan2}
K.~Armanious, C.~Jiang, M.~Fischer, T.~K{\"u}stner, T.~Hepp, K.~Nikolaou,
  S.~Gatidis, and B.~Yang.
\newblock Medgan: Medical image translation using gans.
\newblock {\em Computerized medical imaging and graphics}, 79:101684, 2020.

\bibitem{nonuni}
G.~Batzolis, J.~Stanczuk, C.-B. Schönlieb, and C.~Etmann.
\newblock Non-uniform diffusion models, 2022.

\bibitem{diffi2i1}
G.~Batzolis, J.~Stanczuk, C.-B. Schönlieb, and C.~Etmann.
\newblock Non-uniform diffusion models, 2022.

\bibitem{gan7}
A.~Brock, T.~Lim, J.~Ritchie, and N.~Weston.
\newblock Neural photo editing with introspective adversarial networks.
\newblock In {\em International Conference on Learning Representations}, 2017.

\bibitem{bucher2017generating}
M.~Bucher, S.~Herbin, and F.~Jurie.
\newblock Generating visual representations for zero-shot classification.
\newblock In {\em Proceedings of the IEEE International Conference on Computer
  Vision Workshops}, pages 2666--2673, 2017.

\bibitem{hvae}
A.~L. Caterini, A.~Doucet, and D.~Sejdinovic.
\newblock Hamiltonian variational auto-encoder, 2018.

\bibitem{chaudhuri2020simplified}
U.~Chaudhuri, B.~Banerjee, A.~Bhattacharya, and M.~Datcu.
\newblock A simplified framework for zero-shot cross-modal sketch data
  retrieval.
\newblock In {\em Proceedings of the IEEE/CVF Conference on Computer Vision and
  Pattern Recognition Workshops}, pages 182--183, 2020.

\bibitem{cudac}
J.~Cheng, M.~Grossman, and T.~McKercher.
\newblock {\em Professional CUDA c programming}.
\newblock John Wiley \& Sons, 2014.

\bibitem{adaptiveRealistic}
S.-I. Cheng, Y.-J. Chen, W.-C. Chiu, H.-Y. Tseng, and H.-Y. Lee.
\newblock Adaptively-realistic image generation from stroke and sketch with
  diffusion model, 2022.

\bibitem{diff_cond1}
J.~Choi, S.~Kim, Y.~Jeong, Y.~Gwon, and S.~Yoon.
\newblock Ilvr: Conditioning method for denoising diffusion probabilistic
  models, 2021.

\bibitem{dhariwal2021diffusion}
P.~Dhariwal and A.~Nichol.
\newblock Diffusion models beat gans on image synthesis.
\newblock In M.~Ranzato, A.~Beygelzimer, Y.~Dauphin, P.~Liang, and J.~W.
  Vaughan, editors, {\em Advances in Neural Information Processing Systems},
  volume~34, pages 8780--8794. Curran Associates, Inc., 2021.

\bibitem{ZhangSAR}
Z.~Fu and W.~Zhang.
\newblock Research on image translation between sar and optical imagery.
\newblock {\em ISPRS Annals of the Photogrammetry, Remote Sensing and Spatial
  Information Sciences}, I-7, 07 2012.

\bibitem{pmlr-v162-guo22g}
Y.~Guo, B.~Liu, and D.~Zhao.
\newblock Online continual learning through mutual information maximization.
\newblock In K.~Chaudhuri, S.~Jegelka, L.~Song, C.~Szepesvari, G.~Niu, and
  S.~Sabato, editors, {\em Proceedings of the 39th International Conference on
  Machine Learning}, volume 162 of {\em Proceedings of Machine Learning
  Research}, pages 8109--8126. PMLR, 17--23 Jul 2022.

\bibitem{han2017mr}
X.~Han.
\newblock Mr-based synthetic ct generation using a deep convolutional neural
  network method.
\newblock {\em Medical physics}, 44(4):1408--1419, 2017.

\bibitem{curious1}
A.~Hering, L.~Hansen, T.~C.~W. Mok, A.~C.~S. Chung, H.~Siebert, S.~Häger,
  A.~Lange, S.~Kuckertz, S.~Heldmann, W.~Shao, S.~Vesal, M.~Rusu, G.~Sonn,
  T.~Estienne, M.~Vakalopoulou, L.~Han, Y.~Huang, P.-T. Yap, M.~Brudfors,
  Y.~Balbastre, S.~Joutard, M.~Modat, G.~Lifshitz, D.~Raviv, J.~Lv, Q.~Li,
  V.~Jaouen, D.~Visvikis, C.~Fourcade, M.~Rubeaux, W.~Pan, Z.~Xu, B.~Jian,
  F.~De~Benetti, M.~Wodzinski, N.~Gunnarsson, J.~Sjölund, D.~Grzech, H.~Qiu,
  Z.~Li, A.~Thorley, J.~Duan, C.~Großbröhmer, A.~Hoopes, I.~Reinertsen,
  Y.~Xiao, B.~Landman, Y.~Huo, K.~Murphy, N.~Lessmann, B.~van Ginneken, A.~V.
  Dalca, and M.~P. Heinrich.
\newblock Learn2reg: comprehensive multi-task medical image registration
  challenge, dataset and evaluation in the era of deep learning, 2021.

\bibitem{FID}
M.~Heusel, H.~Ramsauer, T.~Unterthiner, B.~Nessler, and S.~Hochreiter.
\newblock Gans trained by a two time-scale update rule converge to a local nash
  equilibrium.
\newblock In I.~Guyon, U.~V. Luxburg, S.~Bengio, H.~Wallach, R.~Fergus,
  S.~Vishwanathan, and R.~Garnett, editors, {\em Advances in Neural Information
  Processing Systems}, volume~30. Curran Associates, Inc., 2017.

\bibitem{cycleganmt2}
Y.~Hiasa, Y.~Otake, M.~Takao, T.~Matsuoka, K.~Takashima, A.~Carass, J.~L.
  Prince, N.~Sugano, and Y.~Sato.
\newblock Cross-modality image synthesis from unpaired data using cyclegan.
\newblock In {\em International workshop on simulation and synthesis in medical
  imaging}, pages 31--41. Springer, 2018.

\bibitem{hjelm2018learning}
R.~D. Hjelm, A.~Fedorov, S.~Lavoie-Marchildon, K.~Grewal, P.~Bachman,
  A.~Trischler, and Y.~Bengio.
\newblock Learning deep representations by mutual information estimation and
  maximization.
\newblock In {\em International Conference on Learning Representations}, 2019.

\bibitem{DDPM}
J.~Ho, A.~Jain, and P.~Abbeel.
\newblock Denoising diffusion probabilistic models.
\newblock In H.~Larochelle, M.~Ranzato, R.~Hadsell, M.~Balcan, and H.~Lin,
  editors, {\em Advances in Neural Information Processing Systems}, volume~33,
  pages 6840--6851. Curran Associates, Inc., 2020.

\bibitem{classifierfree}
J.~Ho and T.~Salimans.
\newblock Classifier-free diffusion guidance, 2022.

\bibitem{MICrossModal}
T.~Hoang, T.-T. Do, T.~V. Nguyen, and N.-M. Cheung.
\newblock Multimodal mutual information maximization: A novel approach for
  unsupervised deep cross-modal hashing.
\newblock {\em IEEE Transactions on Neural Networks and Learning Systems},
  pages 1--14, 2022.

\bibitem{mr2ct}
T.~Huynh, Y.~Gao, J.~Kang, L.~Wang, P.~Zhang, J.~Lian, and D.~Shen.
\newblock Estimating ct image from mri data using structured random forest and
  auto-context model.
\newblock {\em IEEE Transactions on Medical Imaging}, 35(1):174--183, 2016.

\bibitem{score2}
A.~Hyv{{\"a}}rinen.
\newblock Estimation of non-normalized statistical models by score matching.
\newblock {\em Journal of Machine Learning Research}, 6(24):695--709, 2005.

\bibitem{JiangGAN}
L.~Jiang, M.~Xu, X.~Wang, and L.~Sigal.
\newblock Saliency-guided image translation.
\newblock In {\em Proceedings of the IEEE/CVF Conference on Computer Vision and
  Pattern Recognition (CVPR)}, pages 16509--16518, June 2021.

\bibitem{cycleganmt1}
S.~Joshi, R.~Osuala, C.~Mart{\'\i}n-Isla, V.~M. Campello, C.~Sendra-Balcells,
  K.~Lekadir, and S.~Escalera.
\newblock nn-unet training on cyclegan-translated images for cross-modal domain
  adaptation in biomedical imaging.
\newblock In {\em International MICCAI Brainlesion Workshop}, pages 540--551.
  Springer, 2022.

\bibitem{styleGAN2}
T.~Karras, S.~Laine, M.~Aittala, J.~Hellsten, J.~Lehtinen, and T.~Aila.
\newblock Analyzing and improving the image quality of stylegan, 2019.

\bibitem{diff_con2}
B.~Kawar, M.~Elad, S.~Ermon, and J.~Song.
\newblock Denoising diffusion restoration models, 2022.

\bibitem{kawar2022denoising}
B.~Kawar, M.~Elad, S.~Ermon, and J.~Song.
\newblock Denoising diffusion restoration models.
\newblock {\em arXiv preprint arXiv:2201.11793}, 2022.

\bibitem{gan1}
V.~Kearney, B.~P. Ziemer, A.~Perry, T.~Wang, J.~W. Chan, L.~Ma, O.~Morin, S.~S.
  Yom, and T.~D. Solberg.
\newblock Attention-aware discrimination for mr-to-ct image translation using
  cycle-consistent generative adversarial networks.
\newblock {\em Radiology. Artificial Intelligence}, 2(2), 2020.

\bibitem{zsl}
H.~C. Kuchibhotla, S.~S. Malagi, S.~Chandhok, and V.~N. Balasubramanian.
\newblock Unseen classes at a later time? no problem, 2022.

\bibitem{LevyGen}
D.~Levy, M.~D. Hoffman, and J.~Sohl-Dickstein.
\newblock Generalizing hamiltonian monte carlo with neural networks, 2017.

\bibitem{gan6}
X.~Li, Z.~Du, Y.~Huang, and Z.~Tan.
\newblock A deep translation (gan) based change detection network for optical
  and sar remote sensing images.
\newblock {\em ISPRS Journal of Photogrammetry and Remote Sensing}, 179:14--34,
  2021.

\bibitem{lin2021zstgan}
J.~Lin, Y.~Xia, S.~Liu, S.~Zhao, and Z.~Chen.
\newblock Zstgan: An adversarial approach for unsupervised zero-shot
  image-to-image translation.
\newblock {\em Neurocomputing}, 461:327--335, 2021.

\bibitem{lin2020learning}
K.~Lin, X.~Xu, L.~Gao, Z.~Wang, and H.~T. Shen.
\newblock Learning cross-aligned latent embeddings for zero-shot cross-modal
  retrieval.
\newblock In {\em Proceedings of the AAAI Conference on Artificial
  Intelligence}, volume~34, pages 11515--11522, 2020.

\bibitem{long2017zero}
Y.~Long, L.~Liu, F.~Shen, L.~Shao, and X.~Li.
\newblock Zero-shot learning using synthesised unseen visual data with
  diffusion regularisation.
\newblock {\em IEEE transactions on pattern analysis and machine intelligence},
  40(10):2498--2512, 2017.

\bibitem{AAAIMI}
Y.~Mao, X.~Yan, Q.~Guo, and Y.~Ye.
\newblock Deep mutual information maximin for cross-modal clustering.
\newblock {\em Proceedings of the AAAI Conference on Artificial Intelligence},
  35(10):8893--8901, May 2021.

\bibitem{sdeit}
C.~Meng, Y.~He, Y.~Song, J.~Song, J.~Wu, J.-Y. Zhu, and S.~Ermon.
\newblock {SDE}dit: Guided image synthesis and editing with stochastic
  differential equations.
\newblock In {\em International Conference on Learning Representations}, 2022.

\bibitem{MIDomainAdaptation}
L.~Meng, H.~Su, C.~Lou, and J.~Li.
\newblock Cross-domain mutual information adversarial maximization.
\newblock {\em Engineering Applications of Artificial Intelligence},
  110:104665, 2022.

\bibitem{nie2016estimating}
D.~Nie, X.~Cao, Y.~Gao, L.~Wang, and D.~Shen.
\newblock Estimating ct image from mri data using 3d fully convolutional
  networks.
\newblock In {\em Deep Learning and Data Labeling for Medical Applications},
  pages 170--178. Springer, 2016.

\bibitem{Nyholm2018}
T.~Nyholm, S.~Svensson, S.~Andersson, J.~Jonsson, M.~Sohlin, C.~Gustafsson,
  E.~Kjell{\'e}n, K.~S{\"o}derstr{\"o}m, P.~Albertsson, L.~Blomqvist,
  B.~Zackrisson, L.~E. Olsson, and A.~Gunnlaugsson.
\newblock {MR} and {CT} data with multiobserver delineations of organs in the
  pelvic area-part of the gold atlas project.
\newblock {\em Med. Phys.}, 45(3):1295--1300, Mar. 2018.

\bibitem{nyholm_tufve_2017_583096}
T.~Nyholm, S.~Svensson, S.~Andersson, J.~Jonsson, M.~Sohlin, C.~Gustavsson,
  E.~Kjellen, P.~Albertsson, L.~Blomqvist, B.~Zackrisson, L.~E. Olsson, and
  A.~Gunnlaugsson.
\newblock Gold atlas - male pelvis - gentle radiotherapy, May 2017.

\bibitem{GANDIFF}
M.~Ozbey, S.~U. Dar, H.~A. Bedel, O.~Dalmaz, S.~Ozturk, A.~Gungor, and
  T.~Cukur.
\newblock Unsupervised medical image translation with adversarial diffusion
  models, 2022.

\bibitem{ozdenizci2022}
O.~\"{O}zdenizci and R.~Legenstein.
\newblock Restoring vision in adverse weather conditions with patch-based
  denoising diffusion models.
\newblock {\em arXiv preprint arXiv:2207.14626}, 2022.

\bibitem{palatucci2009zero}
M.~Palatucci, D.~Pomerleau, G.~E. Hinton, and T.~M. Mitchell.
\newblock Zero-shot learning with semantic output codes.
\newblock {\em Advances in neural information processing systems}, 22, 2009.

\bibitem{reviewI2I}
Y.~Pang, J.~Lin, T.~Qin, and Z.~Chen.
\newblock Image-to-image translation: Methods and applications, 2021.

\bibitem{cycleganmt3}
O.~Patashnik, D.~Danon, H.~Zhang, and D.~Cohen-Or.
\newblock Balagan: Cross-modal image translation between imbalanced domains.
\newblock In {\em Proceedings of the IEEE/CVF Conference on Computer Vision and
  Pattern Recognition}, pages 2659--2667, 2021.

\bibitem{SalimansGen}
T.~Salimans, D.~P. Kingma, and M.~Welling.
\newblock Markov chain monte carlo and variational inference: Bridging the gap.
\newblock In {\em ICML}, pages 1218--1226, 2015.

\bibitem{cmtgan4}
A.~Sharma and N.~Jindal.
\newblock Cross-modality breast image translation with improved resolution
  using generative adversarial networks.
\newblock {\em Wireless Personal Communications}, 119(4):2877--2891, Mar. 2021.

\bibitem{SemanticStyleGAN}
Y.~Shi, X.~Yang, Y.~Wan, and X.~Shen.
\newblock Semanticstylegan: Learning compositional generative priors for
  controllable image synthesis and editing, 2021.

\bibitem{cmtgan5}
A.~Sikka, {Skand}, J.~S. Virk, and D.~R. Bathula.
\newblock Mri to pet cross-modality translation using globally and locally
  aware gan (gla-gan) for multi-modal diagnosis of alzheimer's disease, 2021.

\bibitem{Implicit}
J.~Song, C.~Meng, and S.~Ermon.
\newblock Denoising diffusion implicit models, 2020.

\bibitem{SongGen}
J.~Song, S.~Zhao, and S.~Ermon.
\newblock A-nice-mc: Adversarial training for mcmc, 06 2017.

\bibitem{songsde}
Y.~Song, J.~Sohl-Dickstein, D.~P. Kingma, A.~Kumar, S.~Ermon, and B.~Poole.
\newblock Score-based generative modeling through stochastic differential
  equations.
\newblock In {\em International Conference on Learning Representations}, 2021.

\bibitem{cmtgan3}
R.~Toda, A.~Teramoto, M.~Kondo, K.~Imaizumi, K.~Saito, and H.~Fujita.
\newblock Lung cancer {CT} image generation from a free-form sketch using
  style-based pix2pix for data augmentation.
\newblock {\em Scientific Reports}, 12(1), July 2022.

\bibitem{gan5}
H.~Toriya, A.~Dewan, and I.~Kitahara.
\newblock Sar2opt: Image alignment between multi-modal images using generative
  adversarial networks.
\newblock In {\em IGARSS 2019-2019 IEEE International Geoscience and Remote
  Sensing Symposium}, pages 923--926. IEEE, 2019.

\bibitem{van1994grey}
P.~A. Van Den~Elsen, E.-J.~D. Pol, T.~S. Sumanaweera, P.~F. Hemler, S.~Napel,
  and J.~R. Adler.
\newblock Grey value correlation techniques used for automatic matching of ct
  and mr brain and spine images.
\newblock In {\em Visualization in Biomedical Computing 1994}, volume 2359,
  pages 227--237. SPIE, 1994.

\bibitem{score1}
P.~Vincent.
\newblock A connection between score matching and denoising autoencoders.
\newblock {\em Neural Computation}, 23(7):1661--1674, 2011.

\bibitem{gan4}
L.~Wang, B.~Goldluecke, and C.~Anklam.
\newblock L2r gan: Lidar-to-radar translation.
\newblock In {\em Proceedings of the Asian Conference on Computer Vision},
  2020.

\bibitem{wang2018zero}
W.~Wang, Y.~Pu, V.~Verma, K.~Fan, Y.~Zhang, C.~Chen, P.~Rai, and L.~Carin.
\newblock Zero-shot learning via class-conditioned deep generative models.
\newblock In {\em Proceedings of the AAAI Conference on Artificial
  Intelligence}, volume~32, 2018.

\bibitem{wang2019survey}
W.~Wang, V.~W. Zheng, H.~Yu, and C.~Miao.
\newblock A survey of zero-shot learning: Settings, methods, and applications.
\newblock {\em ACM Transactions on Intelligent Systems and Technology (TIST)},
  10(2):1--37, 2019.

\bibitem{flowwang}
Z.~Wang and H.~Delingette.
\newblock Quasi-symplectic langevin variational autoencoder, 2020.

\bibitem{wangfewshot}
Z.~Wang, C.~Vandersteen, C.~Raffaelli, N.~Guevara, F.~Patou, and H.~Delingette.
\newblock One-shot learning for landmarks detection.
\newblock In S.~Engelhardt, I.~Oksuz, D.~Zhu, Y.~Yuan, A.~Mukhopadhyay,
  N.~Heller, S.~X. Huang, H.~Nguyen, R.~Sznitman, and Y.~Xue, editors, {\em
  Deep Generative Models, and Data Augmentation, Labelling, and Imperfections},
  pages 163--172, Cham, 2021. Springer International Publishing.

\bibitem{imagewarp}
G.~Wolberg.
\newblock {\em Digital image warping}.
\newblock Los Alamitos, Calif : IEEE Computer Society Press, US, 1990.

\bibitem{gan3}
J.~M. Wolterink, A.~M. Dinkla, M.~H. Savenije, P.~R. Seevinck, C.~A. van~den
  Berg, and I.~I{\v{s}}gum.
\newblock Deep mr to ct synthesis using unpaired data.
\newblock In {\em International workshop on simulation and synthesis in medical
  imaging}, pages 14--23. Springer, 2017.

\bibitem{gan9}
Z.~Wu, D.~Lischinski, and E.~Shechtman.
\newblock Stylespace analysis: Disentangled controls for stylegan image
  generation.
\newblock In {\em Proceedings of the IEEE/CVF Conference on Computer Vision and
  Pattern Recognition (CVPR)}, pages 12863--12872, June 2021.

\bibitem{8795512}
Y.~Xiao, H.~Rivaz, M.~Chabanas, M.~Fortin, I.~Machado, Y.~Ou, M.~P. Heinrich,
  J.~A. Schnabel, X.~Zhong, A.~Maier, W.~Wein, R.~Shams, S.~Kadoury, D.~Drobny,
  M.~Modat, and I.~Reinertsen.
\newblock Evaluation of mri to ultrasound registration methods for brain shift
  correction: The curious2018 challenge.
\newblock {\em IEEE Transactions on Medical Imaging}, 39(3):777--786, 2020.

\bibitem{cmtgan2}
Q.~Yang, N.~Li, Z.~Zhao, X.~Fan, E.~I.-C. Chang, and Y.~Xu.
\newblock Mri cross-modality neuroimage-to-neuroimage translation, 2018.

\bibitem{cmtgan1}
Q.~Yang, N.~Li, Z.~Zhao, X.~Fan, E.~I.-C. Chang, and Y.~Xu.
\newblock {MRI} cross-modality image-to-image translation.
\newblock {\em Scientific Reports}, 10(1), Feb. 2020.

\bibitem{EnergyGuided}
M.~Zhao, F.~Bao, C.~Li, and J.~Zhu.
\newblock Egsde: Unpaired image-to-image translation via energy-guided
  stochastic differential equations, 2022.

\bibitem{Zheng_2022_CVPR}
X.~Zheng, X.~Fei, L.~Zhang, C.~Wu, F.~Chao, J.~Liu, W.~Zeng, Y.~Tian, and
  R.~Ji.
\newblock Neural architecture search with representation mutual information.
\newblock In {\em Proceedings of the IEEE/CVF Conference on Computer Vision and
  Pattern Recognition (CVPR)}, pages 11912--11921, June 2022.

\bibitem{ganinversion}
J.~Zhu, Y.~Shen, D.~Zhao, and B.~Zhou.
\newblock In-domain gan inversion for real image editing.
\newblock In A.~Vedaldi, H.~Bischof, T.~Brox, and J.-M. Frahm, editors, {\em
  Computer Vision -- ECCV 2020}, pages 592--608, Cham, 2020. Springer
  International Publishing.

\bibitem{CycleGAN2017}
J.-Y. Zhu, T.~Park, P.~Isola, and A.~A. Efros.
\newblock Unpaired image-to-image translation using cycle-consistent
  adversarial networks.
\newblock In {\em Computer Vision (ICCV), 2017 IEEE International Conference
  on}, 2017.

\bibitem{cmtgan6}
H.~Zhuge, B.~Summa, J.~Hamm, and J.~Q. Brown.
\newblock Deep learning 2d and 3d optical sectioning microscopy using
  cross-modality pix2pix cgan image translation.
\newblock {\em Biomed. Opt. Express}, 12(12):7526--7543, Dec 2021.

\end{thebibliography}

\end{document}